\documentclass[conference]{IEEEtran}
\IEEEoverridecommandlockouts
\usepackage{cite}
\usepackage{amsmath,amssymb,amsfonts}
\usepackage{algorithmic}
\usepackage{graphicx}
\usepackage{textcomp}
\usepackage{xcolor}
\usepackage{booktabs}
\usepackage{multirow}
\usepackage{ulem}

\newcommand{\rep}{\ensuremath{^{\dagger}}}
\newcommand{\std}[1]{\textnormal{\tiny$\pm$#1}}
\newcommand{\meanstd}[2]{#1\std{#2}}
\newcommand{\bestpm}[2]{\textbf{#1}\std{#2}}
\newcommand{\secondpm}[2]{\underline{#1\std{#2}}}

\newcommand{\indicator}{\mathbb{I}}
\def\BibTeX{{\rm B\kern-.05em{\sc i\kern-.025em b}\kern-.08em
    T\kern-.1667em\lower.7ex\hbox{E}\kern-.125emX}}
\begin{document}

\title{ProGeo-RC: Prototype-guided Semantic Reliability and Geometric Error Calibration for Semi-supervised Cardiac Image Segmentation
\thanks{This work was supported by the National Key Research and Development Program of China under Grant 2025YFC2708700, in part by Brain-Gain Plan of New Chongqing Foundation under Grant CSTB2024YCJH-KYXM0108, in part by Natural Science Foundation of Chongqing under Grant CSTB2025NSCQ-GPX1037, in part by Natural Science Foundation Innovation and Development Joint Fund Project of Chongqing under Grant CSTB2023NSCQ-LZX0148, in part by Science and Technology Research Program of Chongqing Municipal Education Commission under Grant KJQN202500532 and KJQN202500511.
\protect\newline\hspace*{1em}\textsuperscript{*}Co-first authors contributed equally to this work.
\protect\newline\hspace*{1em}\textsuperscript{\(\dagger\)}Corresponding authors: Wei Hu (wei369.workstation@cqnu.edu.cn); Jianming Hu (hujianming@cqnu.edu.cn).}}

\author{
\IEEEauthorblockN{
Yunfei Zhu\textsuperscript{1*},
Wenfeng Zhang\textsuperscript{1*},
Pan Zeng\textsuperscript{1},
Xin Huang\textsuperscript{1},
Tao Jia\textsuperscript{1},
Weiting Shao\textsuperscript{2},
Wei Hu\textsuperscript{1\(\dagger\)},
Jianming Hu\textsuperscript{3\(\dagger\)}
}
\IEEEauthorblockA{
\textsuperscript{1}College of Computer and Information Science,\\
Research Center of Biomimetic Intelligence and Computational Biomedicine,\\
Chongqing Normal University\\
\textsuperscript{2}School of Educational Sciences, Chongqing Normal University\\
\textsuperscript{3}School of Physics and Electrical Engineering, Chongqing Normal University\\
Chongqing, China
}
}
\maketitle

\begin{abstract}
Semi-supervised medical image segmentation aims to train reliable segmentation models using limited labeled data and abundant unlabeled data. However, existing methods typically rely on prediction confidence or model consistency to estimate pseudo-label reliability, but these criteria are insufficient to characterize both semantic reliability and geometric plausibility. To address this issue, we propose ProGeo-RC, a prototype-guided semantic reliability and geometric error calibration framework for semi-supervised medical image segmentation. ProGeo-RC consists of two key modules: Prototype-guided Semantic Reliability Estimation (PSRE) and Structured Semantic-Geometric Error Calibration (S-SGEC). PSRE uses probabilistic class prototypes to estimate the semantic reliability of unlabeled predictions and calibrate pixel-wise teacher pseudo labels, while S-SGEC introduces signed-distance-field-based geometric constraints to refine ambiguous boundaries and suppress spurious foreground responses outside target structures. Experiments on the ACDC and MVAA-CT datasets show that ProGeo-RC achieves better performance compared with representative semi-supervised segmentation methods. Under the 10\% labeled setting on ACDC, ProGeo-RC improves DSC and Jaccard by 0.37\% and 0.49\% over the corresponding second-best results, while reducing 95HD and ASD by 0.61 and 0.21, respectively. These results indicate that joint semantic reliability estimation and geometric error calibration can provide more reliable pseudo supervision for semi-supervised medical image segmentation.
\end{abstract}


\section{Introduction}
\label{sec:introduction}
Computer-aided diagnosis systems rely on accurate anatomical segmentation for disease assessment, treatment planning, and postoperative follow-up. In cardiac image analysis, reliable segmentation supports ventricular volume quantification, ejection fraction estimation, valve morphology analysis, and personalized treatment planning. Deep learning networks, including U-Net, V-Net, 3D U-Net, and nnU-Net, have achieved strong performance in medical image segmentation~\cite{b1,b2,b3,b4} and cardiac segmentation~\cite{b5,b6,b7,b8,b9}. However, they usually require extensive pixel-level annotations, which are labor-intensive, time-consuming, and clinically demanding. Learning from limited labeled data and abundant unlabeled data therefore remains a key challenge in semi-supervised cardiac image segmentation.

Existing semi-supervised segmentation methods mainly exploit unlabeled data through pseudo-label learning, consistency regularization, teacher-student frameworks, and cross pseudo supervision~\cite{b10,b11,b12,b13}. 
Although these methods have demonstrated the effectiveness of exploiting unlabeled data, their performance largely depends on the reliability of pseudo supervision. Common strategies assess pseudo-label reliability using prediction confidence or model consistency, either by selecting high-confidence regions or by enforcing consistency across perturbations, branches, or teacher-student models~\cite{b11,b12}. However, these criteria are imperfect surrogates for true reliability. High-confidence predictions may still deviate from the semantic distributions of their predicted classes~\cite{b16}, while consistent predictions do not necessarily guarantee geometrically plausible structures~\cite{b17,b18,b19}. Therefore, the key challenge in semi-supervised cardiac image segmentation lies not only in exploiting unlabeled data, but also in identifying reliable pseudo supervision from unlabeled predictions.

This challenge motivates assessing pseudo-supervision reliability from both semantic and geometric perspectives. Semantically, models may produce overconfident errors, where high-confidence pixels do not match the feature distribution of the predicted class~\cite{b16,b20,b21}; agreement between two models also does not necessarily imply feature-space reliability. Geometrically, cardiac structures exhibit anatomical continuity and clear boundaries, making confidence-based filtering insufficient for boundary ambiguity, isolated false positives, and spurious responses outside target structures~\cite{b22,b23,b24}. Although prototypes, intra-class distribution modeling, boundary-aware learning, and distance-based constraints can improve semantic discrimination and structural plausibility, they are usually used as auxiliary regularization or representation learning constraints. These observations suggest that reliable pseudo supervision should jointly consider feature-space semantic consistency and image-space geometric plausibility.

To this end, we propose ProGeo-RC, a prototype-guided semantic reliability and geometric error calibration framework for semi-supervised cardiac image segmentation. Unlike methods relying mainly on prediction confidence or model consistency, ProGeo-RC assesses pseudo-supervision reliability from two complementary aspects: whether pixel features match the predicted class distribution in feature space and whether predicted regions are geometrically plausible in image space. By jointly modeling semantic consistency and geometric plausibility, ProGeo-RC provides more reliable pseudo supervision and reduces the accumulation of erroneous pseudo labels.

Our main contributions are summarized as follows:
\begin{itemize}
    \item We propose ProGeo-RC, a prototype-guided framework for semi-supervised cardiac image segmentation that jointly models pseudo-supervision reliability from complementary semantic and geometric perspectives, thereby mitigating the accumulation of unreliable pseudo labels during training.
    \item We develop Prototype-guided Semantic Reliability Estimation (PSRE) and Structured Semantic-Geometric Error Calibration (S-SGEC) to improve pseudo-supervision quality. PSRE evaluates unlabeled predictions using class-wise probabilistic prototypes, while S-SGEC uses signed distance fields to calibrate boundary ambiguity and suppress structurally implausible foreground responses.
    \item Extensive experiments on the ACDC and MVAA-CT datasets demonstrate that ProGeo-RC consistently outperforms representative semi-supervised segmentation methods under different annotation settings. Ablation studies and qualitative visualizations further verify the effectiveness and generalizability of the proposed framework.
\end{itemize}

\section{Method}
\label{sec:method}

\subsection{Overview}
\label{subsec:overview}

\begin{figure*}[!t]
    \centering
    \includegraphics[width=0.9\textwidth]{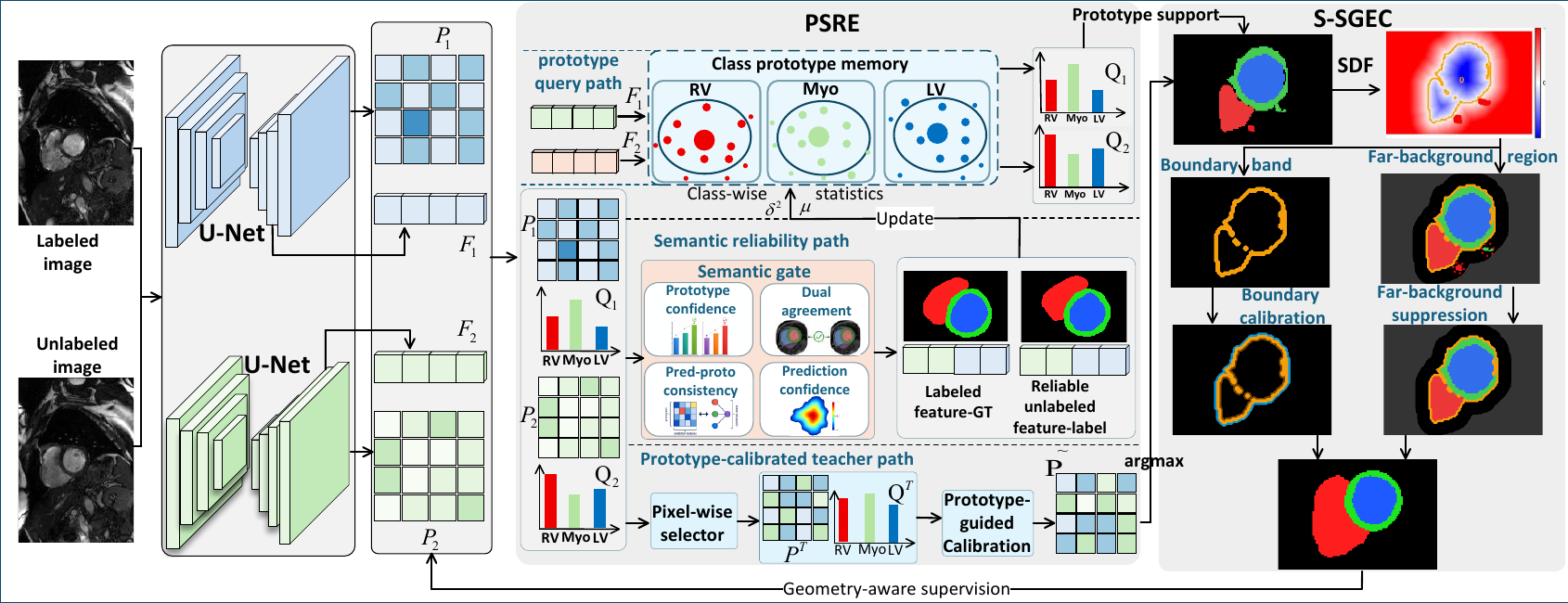}
    \caption{Overall framework of ProGeo-RC. Two student networks generate predictions and features from labeled and unlabeled images. PSRE estimates semantic reliability using probabilistic prototypes and a semantic gate, while S-SGEC applies SDF-based calibration to boundary regions and far-background false positives.}
    \label{fig:framework}
\end{figure*}

We propose ProGeo-RC for semi-supervised cardiac image segmentation under limited annotation. As shown in Fig.~\ref{fig:framework}, the framework contains two student networks with identical architectures but independent parameters. Given a labeled dataset $\mathcal{D}_l=\{(x_i,y_i)\}_{i=1}^{N_l}$ and an unlabeled dataset $\mathcal{D}_u=\{x_i\}_{i=1}^{N_u}$, the $k$-th student network $f_k$ outputs a segmentation probability map $P_k$ and a feature map $F_k$:
\begin{equation}
(P_k,F_k)=f_k(x), \quad k\in\{1,2\}.
\end{equation}
The probability $P_{k,c}(i)$ denotes the prediction of the $k$-th student for class $c$ at position $i$, where $c\in\{0,1,\ldots,C-1\}$. The background class is denoted by $c=0$, and the foreground anatomical classes are denoted by $\mathcal{C}_f=\{1,\ldots,C-1\}$.

ProGeo-RC calibrates unlabeled pseudo-supervision from semantic and geometric perspectives. PSRE models foreground class distributions with probabilistic prototypes to generate a prototype-calibrated teacher pseudo label $\tilde{P}$ and update the prototype memory with reliable feature-label pairs. S-SGEC then constructs signed distance fields from $\tilde{P}$ to regularize ambiguous boundaries and suppress abnormal foreground responses during student optimization.

\subsection{Prototype-guided Semantic Reliability Estimation}
\label{subsec:psre}

PSRE estimates whether an unlabeled foreground prediction is reliable in the feature space. Because softmax confidence and dual-student agreement may retain overconfident errors in locally similar cardiac structures, we introduce foreground class-wise probabilistic prototypes as semantic reliability estimators. Background prototypes are not maintained because background regions are heterogeneous and less suitable for compact prototype modeling.

For the $k$-th student network and foreground class $c\in\mathcal{C}_f$, the probabilistic prototype memory is defined as:
\begin{equation}
\mathcal{M}_{k,c}=\{\mu_{k,c},\sigma^2_{k,c}\},
\quad c\in\mathcal{C}_f .
\end{equation}
The mean $\mu_{k,c}$ and variance $\sigma^2_{k,c}$ describe the class center and intra-class feature dispersion, respectively. During training, $\mathcal{M}_{k,c}$ is updated using labeled feature-ground-truth pairs and reliable unlabeled feature-pseudo-label pairs selected by the semantic reliability gate.

Given an unlabeled pixel feature $\hat{F}_k(i)$, its prototype response to foreground class $c$ is computed as:
\begin{equation}
Z^{\mathrm{proto}}_{k,c}(i)
=
-\frac{1}{d}\sum_{m=1}^{d}
\frac{
\left(\hat{F}_{k,m}(i)-\mu_{k,c,m}\right)^2
}{
\sigma^2_{k,c,m}+\epsilon
},
\quad c\in\mathcal{C}_f ,
\end{equation}
where $d$ is the feature dimension and $\epsilon$ is a numerical stabilizer. A larger $Z^{\mathrm{proto}}_{k,c}(i)$ indicates stronger consistency between the pixel feature and the prototype distribution of class $c$. The responses are normalized over foreground classes:
\begin{equation}
Q_{k,c}(i)
=
\frac{
\exp\left(Z^{\mathrm{proto}}_{k,c}(i)\right)
}{
\sum_{r\in\mathcal{C}_f}
\exp\left(Z^{\mathrm{proto}}_{k,r}(i)\right)
},
\quad c\in\mathcal{C}_f .
\end{equation}

$Q_{k,c}(i)$ denotes the prototype support for foreground class $c$, while the background probability is retained from the segmentation prediction.

We denote the predicted class and prediction confidence of the $k$-th student as:
\begin{equation}
\hat{y}_k(i)=\operatorname*{argmax}_{c}P_{k,c}(i),
\quad
p_k(i)=\max_c P_{k,c}(i),
\end{equation}
and define the prototype-predicted foreground class and prototype confidence as:
\begin{equation}
\hat{y}^{\mathrm{proto}}_k(i)
=
\operatorname*{argmax}_{c\in\mathcal{C}_f}Q_{k,c}(i),
\quad
q_k(i)=\max_{c\in\mathcal{C}_f} Q_{k,c}(i).
\end{equation}
The semantic reliability gate is formulated as:
\begin{equation}
\begin{aligned}
R^k_{\mathrm{sem}}(i)
=&\ \indicator[\hat{y}_k(i)\in\mathcal{C}_f]
   \indicator[p_k(i)>\tau_p]  \\
&\cdot \indicator[\hat{y}_k(i)=\hat{y}^{\mathrm{proto}}_k(i)]
   \indicator[q_k(i)>\tau_p]  \\
&\cdot \indicator[\hat{y}_1(i)=\hat{y}_2(i)] .
\end{aligned}
\end{equation}
The indicator function is denoted by $\indicator[\cdot]$, and $\tau_p$ is the reliability threshold. This gate jointly evaluates foreground prediction, prediction confidence, prediction-prototype consistency, prototype confidence, and dual-student agreement. Pixels satisfying the gate are treated as reliable foreground feature-label pairs for prototype memory update.

For each foreground class $c\in\mathcal{C}_f$, the prototype update set is constructed as:
\begin{equation}
\begin{aligned}
\mathcal{S}^{l}_{k,c}
&=
\{\hat{F}_k(i)\mid (x,y)\in\mathcal{D}_l,\ y(i)=c\},\\
\mathcal{S}^{u}_{k,c}
&=
\{\hat{F}_k(i)\mid x\in\mathcal{D}_u,\ 
R^k_{\mathrm{sem}}(i)=1,\ \hat{y}_k(i)=c\},\\
\mathcal{S}_{k,c}
&=
\mathcal{S}^{l}_{k,c}\cup \mathcal{S}^{u}_{k,c}.
\end{aligned}
\end{equation}
When $\mathcal{S}_{k,c}$ is non-empty, its batch-wise mean and variance are computed by:
\begin{equation}
\begin{aligned}
\bar{\mu}_{k,c}
&=
\frac{1}{|\mathcal{S}_{k,c}|}
\sum_{\hat{F}_k(i)\in\mathcal{S}_{k,c}}
\hat{F}_k(i),\\
\bar{\sigma}^{2}_{k,c}
&=
\frac{1}{|\mathcal{S}_{k,c}|}
\sum_{\hat{F}_k(i)\in\mathcal{S}_{k,c}}
\left(\hat{F}_k(i)-\bar{\mu}_{k,c}\right)^2 .
\end{aligned}
\end{equation}
The prototype memory is updated using exponential moving average:
\begin{equation}
\begin{aligned}
\mu_{k,c}
&\leftarrow
\eta \mu_{k,c}
+
(1-\eta)\bar{\mu}_{k,c},\\
\sigma^{2}_{k,c}
&\leftarrow
\eta \sigma^{2}_{k,c}
+
(1-\eta)\bar{\sigma}^{2}_{k,c},
\end{aligned}
\end{equation}
where $\eta$ denotes the EMA momentum. If no reliable sample of class $c$ is available in the current mini-batch, $\mathcal{M}_{k,c}$ remains unchanged. This reduces the risk of updating the prototype memory with unreliable pseudo labels.

PSRE further constructs a prototype-calibrated soft teacher pseudo label from the two student predictions. Instead of introducing an additional teacher network, we use a pixel-wise selector to choose the student branch with higher prediction confidence at each unlabeled pixel:
\begin{equation}
\begin{aligned}
s(i) &= \operatorname*{argmax}_{k\in\{1,2\}} p_k(i),\\
P^T_c(i) &= P_{s(i),c}(i), \quad c\in\{0,1,\ldots,C-1\},\\
Q^T_c(i) &= Q_{s(i),c}(i), \quad c\in\mathcal{C}_f .
\end{aligned}
\end{equation}
The selected branch index is $s(i)$. Thus, $P^T_c(i)$ denotes the pixel-wise teacher prediction probability, and $Q^T_c(i)$ denotes the prototype probability queried from the same selected branch. We define:
\begin{equation}
\begin{aligned}
\hat{y}^{T}(i) &= \operatorname*{argmax}_{c} P^T_c(i), \\
\hat{y}^{\mathrm{proto},T}(i)
&=
\operatorname*{argmax}_{c\in\mathcal{C}_f} Q^T_c(i), \\
q^T(i) &= \max_{c\in\mathcal{C}_f} Q^T_c(i).
\end{aligned}
\end{equation}
The prototype calibration indicator is:
\begin{equation}
m(i)=
\indicator[\hat{y}^{T}(i)\in\mathcal{C}_f]
\indicator[\hat{y}^{T}(i)=\hat{y}^{\mathrm{proto},T}(i)]
\indicator[q^T(i)>\tau_p].
\end{equation}

When the selected teacher prediction passes the foreground prototype-consistency criterion, its foreground probabilities are calibrated by prototype responses. To preserve the classifier-derived background probability, we define the foreground probability mass as:
\begin{equation}
\pi^T(i)=\sum_{c\in\mathcal{C}_f}P^T_c(i)=1-P^T_0(i).
\end{equation}
The prototype-calibrated teacher is given by:
\begin{equation}
\tilde{P}_0(i)=P^T_0(i),
\end{equation}
and for $c\in\mathcal{C}_f$,
\begin{equation}
\begin{aligned}
\tilde{P}_c(i)
=&\ (1-m(i))P^T_c(i) \\
&+m(i)\big[(1-\lambda)P^T_c(i)
+\lambda \pi^T(i)Q^T_c(i)\big].
\end{aligned}
\end{equation}
The coefficient $\lambda$ controls the contribution of the prototype probability. Since $Q^T$ is normalized only over foreground classes, multiplication by $\pi^T(i)$ preserves the foreground probability mass and keeps $\tilde{P}$ a valid class distribution. The output $\tilde{P}(i)=\{\tilde{P}_c(i)\}_{c=0}^{C-1}$ is used for subsequent unlabeled supervision. The prototype memory is not updated by $\tilde{P}$, but only by the reliable feature-label pairs selected by $R^k_{\mathrm{sem}}$.

\subsection{Structured Semantic-Geometric Error Calibration}
\label{subsec:ssgec}

Semantic reliability alone does not ensure spatial plausibility, as pseudo-label errors often occur near ambiguous boundaries or outside target structures. Motivated by boundary- and distance-based constraints~\cite{b22,b23,b24}, S-SGEC constructs geometric supervision from the prototype-calibrated teacher prediction $\tilde{P}$.

The soft teacher prediction is first converted into a hard pseudo label:
\begin{equation}
\tilde{Y}(i)=\operatorname*{argmax}_{c}\tilde{P}_{c}(i).
\end{equation}
For each foreground class $c\in\mathcal{C}_f$, a signed distance field (SDF) is constructed as:
\begin{equation}
S_c=\operatorname{SDF}\big(\indicator[\tilde{Y}=c]\big).
\end{equation}
 $S_c(i)<0$, $S_c(i)\approx0$, and $S_c(i)>0$ indicate inside, boundary-adjacent, and outside regions of class $c$, respectively. If class $c$ is absent from $\tilde{Y}$, its geometric weight is set to zero.

To obtain scale-consistent responses, the SDF is normalized by:
\begin{equation}
\bar{S}_c(i)=
\frac{S_c(i)}{\max_j |S_c(j)|+\epsilon}.
\end{equation}
The boundary-region and far-background-region weights are defined as:
\begin{equation}
W^{bd}_c(i)=\big(1-|\bar{S}_c(i)|\big)^2,
\quad
W^{far}_c(i)=\big[\max(\bar{S}_c(i),0)\big]^2 .
\end{equation}
$W^{bd}_c(i)$ emphasizes boundary-adjacent regions, while $W^{far}_c(i)$ highlights outside regions with potential abnormal foreground responses.

Let $\Omega$ denote all pixel or voxel positions in the current image or patch. We use the maximum teacher response as a continuous reliability weight:
\begin{equation}
\rho(i)=\max_c \tilde{P}_c(i).
\end{equation}
The boundary calibration loss is:
\begin{equation}
\mathcal{L}^{k}_{bd}
=
\frac{
\sum\limits_{i\in\Omega,\,c\in\mathcal{C}_f}
\left|P_{k,c}(i)-\tilde{P}_{c}(i)\right|
W^{bd}_c(i)\rho(i)Q_{k,c}(i)
}{
\sum\limits_{i\in\Omega,\,c\in\mathcal{C}_f}
W^{bd}_c(i)\rho(i)Q_{k,c}(i)+\epsilon
}.
\end{equation}
This term aligns the student prediction with the prototype-calibrated teacher in boundary-sensitive regions.

To suppress abnormal foreground responses, we define a class-specific prototype conflict indicator:
\begin{equation}
\begin{aligned}
C_k(i,c)
=&\ \indicator[c=\tilde{Y}(i)]
\indicator[\hat{y}^{\mathrm{proto}}_k(i)\ne \tilde{Y}(i)] \\
&\cdot \indicator[q_k(i)>\tau_p],
\quad c\in\mathcal{C}_f .
\end{aligned}
\end{equation}
This indicator acts only on the teacher-indicated class to avoid indiscriminate foreground suppression. The abnormal foreground weight is:
\begin{equation}
A_k(i,c)
=
W^{far}_c(i)\big(1-Q_{k,c}(i)\big)
+
C_k(i,c),
\quad c\in\mathcal{C}_f .
\end{equation}
The abnormal foreground suppression loss is:
\begin{equation}
\mathcal{L}^{k}_{abn}
=
\frac{
\sum\limits_{i\in\Omega,\,c\in\mathcal{C}_f}
P^{2}_{k,c}(i)A_k(i,c)\rho(i)
}{
\sum\limits_{i\in\Omega,\,c\in\mathcal{C}_f}
A_k(i,c)\rho(i)+\epsilon
}.
\end{equation}
This term penalizes foreground responses in structure-external and class-conflicting regions. The geometric calibration objective for the $k$-th student is:
\begin{equation}
\mathcal{L}^{k}_{geo}
=
\mathcal{L}^{k}_{bd}
+
\mathcal{L}^{k}_{abn}.
\end{equation}
Thus, $\mathcal{L}^{k}_{bd}$ calibrates boundary consistency, while $\mathcal{L}^{k}_{abn}$ suppresses abnormal foreground responses based on the semantically calibrated teacher pseudo label from PSRE.
\subsection{Training Objective}
\label{subsec:training_objective}

The overall objective consists of supervised learning, prototype-calibrated pseudo supervision, and geometric calibration. For a labeled sample $(x^l,Y^l)\in\mathcal{D}_l$, both student networks are optimized by:
\begin{equation}
\mathcal{L}_{sup}
=
\frac{1}{2}\sum_{k=1}^{2}
\left[
\mathcal{L}_{ce}(P^l_k,Y^l)
+
\mathcal{L}_{dice}(P^l_k,Y^l)
\right],
\end{equation}
where $(P^l_k,F^l_k)=f_k(x^l)$, and $\mathcal{L}_{ce}$ and $\mathcal{L}_{dice}$ denote cross-entropy and Dice losses, respectively.

For unlabeled data, the prototype-calibrated soft teacher pseudo label $\tilde{P}$ supervises both student networks. To avoid gradients flowing through the generated pseudo target, $\tilde{P}$ is treated as a detached target. With $\operatorname{sg}[\cdot]$ denoting stop-gradient, we define:
\begin{equation}
\tilde{P}^{\mathrm{sg}}(i)=\operatorname{sg}\left[\tilde{P}(i)\right].
\end{equation}
The unlabeled pseudo-supervision loss is:
\begin{equation}
\mathcal{L}_{u}
=
\frac{1}{2}\sum_{k=1}^{2}
\frac{1}{|\Omega|}
\sum_{i\in\Omega}
\mathcal{H}\big(\tilde{P}^{\mathrm{sg}}(i),P_k(i)\big),
\end{equation}
where $\mathcal{H}(\cdot,\cdot)$ is the soft cross-entropy:
\begin{equation}
\mathcal{H}\big(\tilde{P}^{\mathrm{sg}}(i),P_k(i)\big)
=
-\sum_{c=0}^{C-1}
\tilde{P}^{\mathrm{sg}}_{c}(i)
\log\left(P_{k,c}(i)+\epsilon\right).
\end{equation}
The geometric calibration loss is averaged over the two student networks:
\begin{equation}
\mathcal{L}_{geo}
=
\frac{1}{2}\sum_{k=1}^{2}
\mathcal{L}^{k}_{geo}.
\end{equation}
The final training objective is:
\begin{equation}
\mathcal{L}
=
\mathcal{L}_{sup}
+
\lambda_{u}\mathcal{L}_{u}
+
\lambda_{g}\mathcal{L}_{geo}.
\end{equation}
The coefficients $\lambda_u$ and $\lambda_g$ balance prototype-calibrated pseudo supervision and geometry-aware calibration. In our implementation, they are treated as learnable loss-balancing weights and optimized jointly with the segmentation networks.

\section{Experiments}
\label{sec:experiments}

\subsection{Datasets and Evaluation Metrics}
\label{subsec:datasets_metrics}

We evaluated ProGeo-RC on ACDC~\cite{b5} and MVAA-CT~\cite{b15}. ACDC is a cardiac MRI benchmark from the MICCAI 2017 Automatic Cardiac Diagnosis Challenge, containing 100 cine MR cases annotated for RV, Myo, and LV. Following common semi-supervised settings, we split ACDC into 70 training, 10 validation, and 20 test cases, and used 10\% or 20\% of the training cases as labeled data. MVAA-CT is from the MICCAI 2026 Mitral Valve Anatomy Analysis Using Multimodal Imaging Data Challenge, containing 27 labeled mitral valve CT volumes and 1,040 official unlabeled volumes. Since the official validation and test sets were unavailable, we fixed 7 labeled cases as the held-out test split and used the remaining 20 as the training pool. For each low-annotation setting, 2 or 4 labeled cases were sampled five times, with the remaining training-pool cases treated as unlabeled data. Together with the official unlabeled volumes, the 2-labeled and 4-labeled settings contained 1,058 and 1,056 unlabeled cases, respectively, and results are reported as the mean and standard deviation over five samplings.

We evaluated segmentation performance using DSC, Jaccard, 95\% Hausdorff Distance (95HD), and Average Surface Distance (ASD). DSC and Jaccard quantify regional overlap, while 95HD and ASD measure boundary distance errors.

\subsection{Implementation Details}
\label{subsec:implementation}

All experiments were implemented in PyTorch on a single NVIDIA GeForce RTX 3090 GPU. For ACDC, images were cropped to $256\times256$, with each mini-batch containing 8 labeled and 8 unlabeled images. For MVAA-CT, 3D patches of size $64\times128\times128$ were used, with 2 labeled and 2 unlabeled patches per mini-batch. Training lasted 30,000 iterations for ACDC and 15,000 for MVAA-CT. All reproduced baselines and ProGeo-RC were optimized with SGD using an initial learning rate of $1\times10^{-4}$, momentum of 0.9, and weight decay of $1\times10^{-4}$. The EMA momentum for prototype memory was 0.99, $\lambda_u$ and $\lambda_g$ were learnable non-negative weights, and $\epsilon$ was fixed to $10^{-6}$.

\begin{table}[!t]
\centering
\caption{Comparison on the ACDC test set. Best and second-best semi-supervised results within each labeled split are shown in bold and underlined, respectively. \textsuperscript{\dag} indicates results reproduced by us.}
\label{tab:acdc_main}
\tiny
\setlength{\tabcolsep}{1.2pt}
\renewcommand{\arraystretch}{0.8}
\resizebox{\columnwidth}{!}{%
\begin{tabular}{l|cc|cccc}
\toprule
Method & Lab. & Unlab. & DSC$\uparrow$ & Jaccard$\uparrow$ & 95HD$\downarrow$ & ASD$\downarrow$ \\
\midrule
\multirow{3}{*}{U-Net\textsuperscript{\dag}~\cite{b1}}
& 7(10\%)  & 0 & 77.07 & 65.14 & 8.47 & 2.65 \\
& 14(20\%) & 0 & 84.12 & 74.21 & 6.46 & 2.03 \\
& 70(All)  & 0 & 91.47 & 84.68 & 1.40 & 0.34 \\
\midrule
CPS\textsuperscript{\dag}~\cite{b12}      & \multirow{10}{*}{7(10\%)}  & \multirow{10}{*}{63(90\%)} & 86.74 & 77.43 & 6.64 & 1.80 \\
MC-Net\textsuperscript{\dag}~\cite{b26}   &                            &                            & 87.17 & 77.97 & 8.09 & 2.06 \\
MC-Net+\textsuperscript{\dag}~\cite{b27}  &                            &                            & 86.31 & 76.71 & 8.23 & 2.20 \\
BCP\textsuperscript{\dag}~\cite{b13}      &                            &                            & 88.23 & 79.71 & 4.54 & 1.41 \\
DiffRect\textsuperscript{\dag}~\cite{b28} &                            &                            & \underline{89.53} & \underline{81.57} & 4.33 & 1.21 \\
MLRP\textsuperscript{\dag}~\cite{b17}     &                            &                            & 87.99 & 79.28 & 3.28 & 0.90 \\
W2SPC~\cite{b29}                          &                            &                            & 88.92 & 79.65 & 4.32 & 1.19 \\
BoCLIS~\cite{b24}                         &                            &                            & 88.96 & 79.72 & 4.23 & 1.21 \\
AdaMix\textsuperscript{\dag}~\cite{b30}   &                            &                            & 89.02 & 80.57 & \underline{2.15} & \underline{0.78} \\
\textbf{Ours}                             &                            &                            & \textbf{89.90} & \textbf{82.06} & \textbf{1.54} & \textbf{0.57} \\
\midrule
CPS\textsuperscript{\dag}~\cite{b12}      & \multirow{10}{*}{14(20\%)} & \multirow{10}{*}{56(80\%)} & 88.48 & 80.28 & 5.47 & 1.59 \\
MC-Net\textsuperscript{\dag}~\cite{b26}   &                            &                            & 87.62 & 78.92 & 5.55 & 1.73 \\
MC-Net+\textsuperscript{\dag}~\cite{b27}  &                            &                            & 87.94 & 79.50 & 5.77 & 1.66 \\
BCP\textsuperscript{\dag}~\cite{b13}      &                            &                            & 89.20 & 81.13 & 4.80 & 1.56 \\
DiffRect\textsuperscript{\dag}~\cite{b28} &                            &                            & 89.92 & \underline{82.27} & 3.44 & 0.93 \\
MLRP\textsuperscript{\dag}~\cite{b17}     &                            &                            & 88.76 & 80.64 & 4.49 & 1.16 \\
W2SPC~\cite{b29}                          &                            &                            & \underline{90.23} & 81.24 & 4.01 & 1.06 \\
BoCLIS~\cite{b24}                         &                            &                            & 90.10 & 81.11 & 4.14 & 1.06 \\
AdaMix\textsuperscript{\dag}~\cite{b30}   &                            &                            & 89.52 & 81.29 & \underline{1.97} & \underline{0.64} \\
\textbf{Ours}                             &                            &                            & \textbf{90.72} & \textbf{83.45} & \textbf{1.42} & \textbf{0.45} \\
\bottomrule
\end{tabular}%
}
\end{table}

\subsection{Comparison with State-of-the-art Methods}
\label{subsec:comparison}
We compared ProGeo-RC with representative semi-supervised methods, including CPS~\cite{b12}, MC-Net/MC-Net+~\cite{b26,b27}, BCP~\cite{b13}, DiffRect~\cite{b28}, MLRP~\cite{b17}, W2SPC~\cite{b29}, BoCLIS~\cite{b24}, AdaMix~\cite{b30}, CAML~\cite{b31}, LeFeD~\cite{b14}, SDCL~\cite{b32}, and PICK~\cite{b33}. U-Net~\cite{b1} and 3D U-Net~\cite{b3} served as supervised baselines for ACDC and MVAA-CT, respectively.

The results on ACDC and MVAA-CT are reported in Tables~\ref{tab:acdc_main} and~\ref{tab:mvaa_ct_main}, respectively. On ACDC, ProGeo-RC achieved 89.90\% DSC, 82.06\% Jaccard, 1.54 95HD, and 0.57 ASD with 10\% labeled data, improving the corresponding second-best results by 0.37\%, 0.49\%, 0.61, and 0.21, respectively. With 20\% labeled data, it also ranked first across all metrics, reaching 90.72\% DSC, 83.45\% Jaccard, 1.42 95HD, and 0.45 ASD. On the held-out MVAA-CT test split, results are reported over five random labeled-set samplings. ProGeo-RC achieved 75.80\% DSC, 62.76\% Jaccard, 2.25 95HD, and 0.67 ASD in the 2-labeled setting, and 82.21\% DSC, 69.89\% Jaccard, 1.35 95HD, and 0.43 ASD in the 4-labeled setting, ranking first across all metrics in both cases. It also obtained the smallest standard deviations among semi-supervised methods, indicating stable performance across labeled-set samplings. These results demonstrate improved regional completeness and boundary localization on ACDC, as well as effective CT-based mitral valve region segmentation on MVAA-CT.

\begin{table}[!t]
\centering

\caption{Comparison on the held-out MVAA-CT test split. Best and second-best semi-supervised results within each labeled split are shown in bold and underlined, respectively. \rep indicates results reproduced by us.}
\label{tab:mvaa_ct_main}
\scriptsize
\setlength{\tabcolsep}{1.2pt}
\renewcommand{\arraystretch}{0.8}
\resizebox{\columnwidth}{!}{%
\begin{tabular}{l|cc|cccc}
\toprule
Method & Lab. & Unlab. & DSC$\uparrow$ & Jaccard$\uparrow$ & 95HD$\downarrow$ & ASD$\downarrow$ \\
\midrule
\multirow{2}{*}{3D U-Net\rep~\cite{b3}}
& 2  & 0 & \meanstd{74.55}{3.83} & \meanstd{60.13}{4.90} & \meanstd{4.56}{2.10} & \meanstd{1.18}{0.44} \\
& 4  & 0 & \meanstd{81.06}{0.93} & \meanstd{68.34}{1.27} & \meanstd{1.83}{0.43} & \meanstd{0.59}{0.14} \\
\midrule
CPS\rep~\cite{b12}      & \multirow{9}{*}{2} & \multirow{9}{*}{1058} & \meanstd{72.78}{3.56} & \meanstd{58.74}{3.71} & \meanstd{3.42}{1.34} & \meanstd{1.05}{0.56} \\
MC-Net+\rep~\cite{b27}  &                       &                         & \meanstd{73.28}{2.21} & \meanstd{58.92}{3.36} & \meanstd{3.23}{1.02} & \secondpm{0.74}{0.25} \\
CAML\rep~\cite{b31}     &                       &                         & \meanstd{72.33}{3.17} & \meanstd{57.25}{4.02} & \meanstd{3.79}{1.47} & \meanstd{0.80}{0.42} \\
LeFeD\rep~\cite{b14}    &                       &                         & \secondpm{75.48}{4.33} & \secondpm{61.14}{5.35} & \meanstd{3.23}{1.31} & \meanstd{0.94}{0.33} \\
MLRP\rep~\cite{b17}     &                       &                         & \meanstd{71.46}{3.52} & \meanstd{56.94}{4.13} & \meanstd{4.26}{1.67} & \meanstd{1.15}{0.94} \\
SDCL\rep~\cite{b32}     &                       &                         & \meanstd{74.04}{3.05} & \meanstd{60.45}{3.43} & \secondpm{3.06}{1.21} & \meanstd{0.78}{0.60} \\
PICK\rep~\cite{b33}     &                       &                         & \meanstd{73.55}{5.21} & \meanstd{59.16}{5.53} & \meanstd{3.42}{1.24} & \meanstd{0.77}{0.18} \\
AdaMix\rep~\cite{b30}   &                       &                         & \meanstd{72.59}{3.02} & \meanstd{57.26}{3.67} & \meanstd{4.11}{2.48} & \meanstd{1.52}{0.49} \\
\textbf{Ours}           &                       &                         & \bestpm{75.80}{1.80} & \bestpm{62.76}{2.33} & \bestpm{2.25}{0.60} & \bestpm{0.67}{0.11} \\
\midrule
CPS\rep~\cite{b12}      & \multirow{9}{*}{4} & \multirow{9}{*}{1056} & \meanstd{79.83}{1.50} & \meanstd{65.74}{1.65} & \meanstd{1.97}{0.54} & \meanstd{0.67}{0.31} \\
MC-Net+\rep~\cite{b27}  &                       &                         & \meanstd{80.32}{1.44} & \meanstd{66.75}{1.56} & \meanstd{1.98}{0.52} & \meanstd{0.65}{0.28} \\
CAML\rep~\cite{b31}     &                       &                         & \meanstd{79.87}{1.22} & \meanstd{66.43}{1.37} & \meanstd{2.02}{0.58} & \meanstd{0.72}{0.26} \\
LeFeD\rep~\cite{b14}    &                       &                         & \secondpm{81.27}{1.19} & \secondpm{68.73}{1.59} & \meanstd{1.62}{0.47} & \secondpm{0.51}{0.19} \\
MLRP\rep~\cite{b17}     &                       &                         & \meanstd{80.99}{1.78} & \meanstd{68.28}{2.44} & \meanstd{1.66}{0.56} & \meanstd{0.53}{0.22} \\
SDCL\rep~\cite{b32}     &                       &                         & \meanstd{80.05}{1.54} & \meanstd{67.30}{1.56} & \meanstd{1.67}{0.66} & \meanstd{0.65}{0.27} \\
PICK\rep~\cite{b33}     &                       &                         & \meanstd{80.08}{1.13} & \meanstd{67.10}{1.36} & \meanstd{2.04}{0.62} & \meanstd{0.65}{0.20} \\
AdaMix\rep~\cite{b30}   &                       &                         & \meanstd{81.04}{1.60} & \meanstd{68.71}{2.26} & \secondpm{1.44}{0.43} & \meanstd{0.55}{0.23} \\
\textbf{Ours}           &                       &                         & \bestpm{82.21}{1.01} & \bestpm{69.89}{1.20} & \bestpm{1.35}{0.41} & \bestpm{0.43}{0.17} \\
\bottomrule
\end{tabular}%
}
\end{table}

\subsection{Ablation Study}
\label{subsec:ablation}

We conducted ablation experiments on the ACDC test set to evaluate the contributions of PSRE and S-SGEC, as reported in Table~\ref{tab:ablation}. Adding PSRE improved DSC and Jaccard to 88.77\% and 80.46\%, while reducing 95HD and ASD to 3.40 and 1.01, indicating improved pseudo-label reliability. Adding S-SGEC also yielded clear gains, especially in boundary-related metrics, reducing 95HD and ASD to 3.20 and 0.76. Combining both modules achieved the best performance across all metrics, showing that semantic reliability estimation and geometric error calibration act complementarily. We also analyzed the reliability threshold $\tau_p$ and teacher calibration weight $\lambda$ in Table~\ref{tab:param_sensitivity}. ProGeo-RC remained stable when $\tau_p$ varied from 0.6 to 0.8 and $\lambda$ from 0.1 to 0.3, with the best overall balance obtained at $\tau_p=0.7$ and $\lambda=0.2$.

\begin{table}[!t]
\centering
\caption{Ablation study of the proposed components on the ACDC dataset.}
\label{tab:ablation}
\scriptsize
\setlength{\tabcolsep}{3.2pt}
\renewcommand{\arraystretch}{1.05}
\resizebox{\columnwidth}{!}{%
\begin{tabular}{lcc|cccc}
\toprule
Method & PSRE & S-SGEC & DSC$\uparrow$ & Jaccard$\uparrow$ & 95HD$\downarrow$ & ASD$\downarrow$ \\
\midrule
Baseline & -- & -- & 86.47 & 76.52 & 6.72 & 2.05 \\
Baseline + PSRE & \checkmark & -- & 88.77 & 80.46 & 3.40 & 1.01 \\
Baseline + S-SGEC & -- & \checkmark & 88.69 & 80.29 & 3.20 & 0.76 \\
ProGeo-RC & \checkmark & \checkmark & \textbf{89.90} & \textbf{82.06} & \textbf{1.54} & \textbf{0.57} \\
\bottomrule
\end{tabular}%
}
\end{table}

\begin{table}[!t]
\centering
\caption{Sensitivity analysis of reliability threshold and teacher calibration weight on the ACDC dataset.}
\label{tab:param_sensitivity}
\scriptsize
\setlength{\tabcolsep}{4.2pt}
\renewcommand{\arraystretch}{1.05}
\resizebox{\columnwidth}{!}{%
\begin{tabular}{lc|cccc}
\toprule
Setting & Value & DSC$\uparrow$ & Jaccard$\uparrow$ & 95HD$\downarrow$ & ASD$\downarrow$ \\
\midrule
\multirow{3}{*}{Reliability threshold $\tau_p$}
& 0.6 & 89.82 & 82.01 & 1.57 & 0.60 \\
& 0.7 & \textbf{89.90} & \textbf{82.06} & 1.54 & \textbf{0.57} \\
& 0.8 & 89.88 & 82.04 & \textbf{1.53} & 0.61 \\
\midrule
\multirow{3}{*}{Calibration weight $\lambda$}
& 0.1 & \textbf{89.91} & 82.01 & 1.60 & 0.59 \\
& 0.2 & 89.90 & \textbf{82.06} & \textbf{1.54} & \textbf{0.57} \\
& 0.3 & 89.78 & 81.97 & 1.66 & 0.61 \\
\bottomrule
\end{tabular}%
}
\end{table}

\subsection{Analysis of Visualization}
\label{subsec:visualization}

\begin{figure}[!t]
    \centering
    \includegraphics[width=0.85\columnwidth]{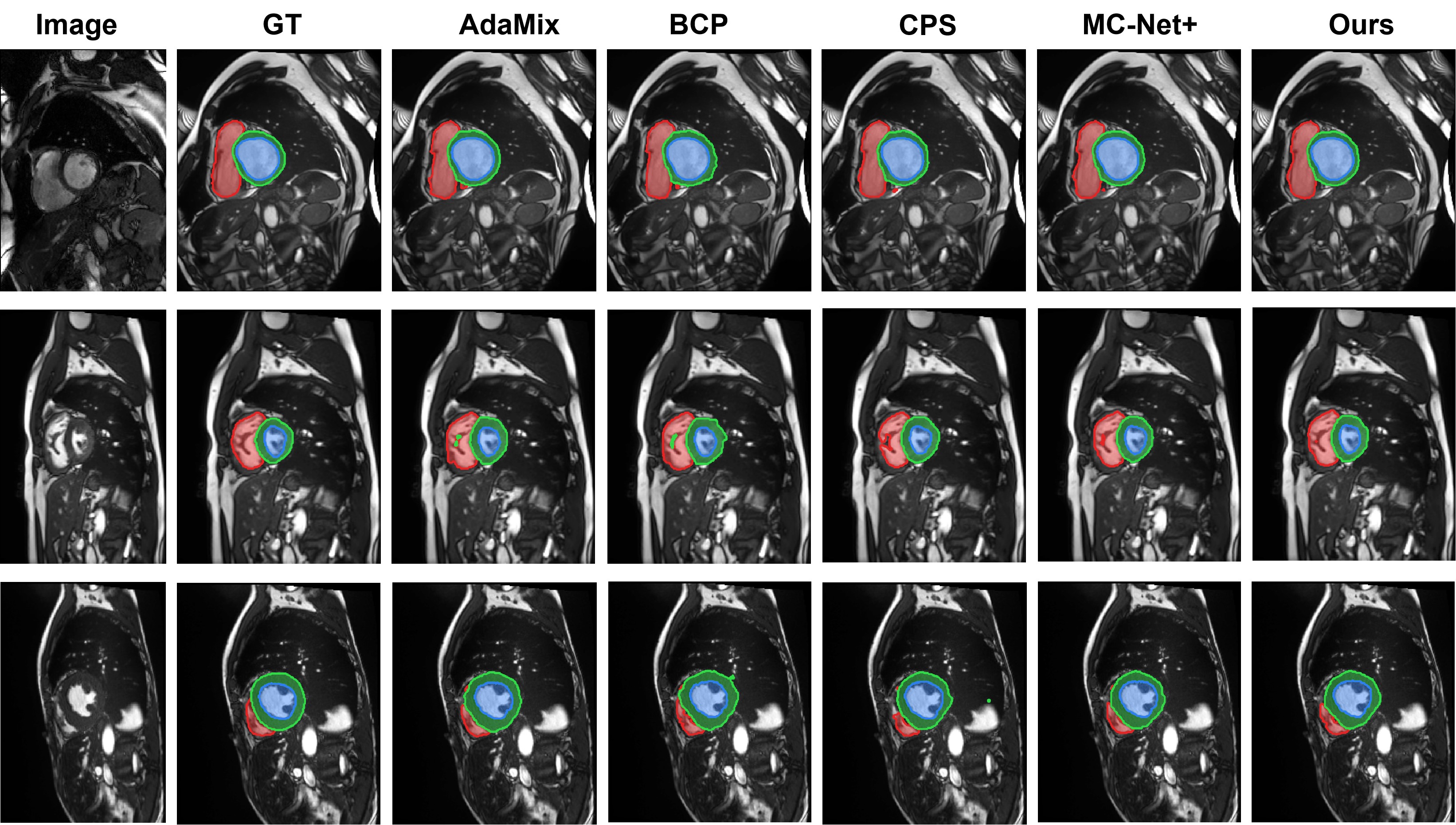}
    \caption{Visual comparison on representative ACDC test slices. Red, green, and blue denote RV, Myo, and LV, respectively.}
    \label{fig:acdc_visual_comparison}
\end{figure}

\begin{figure}[!t]
    \centering
    \includegraphics[width=0.85\columnwidth]{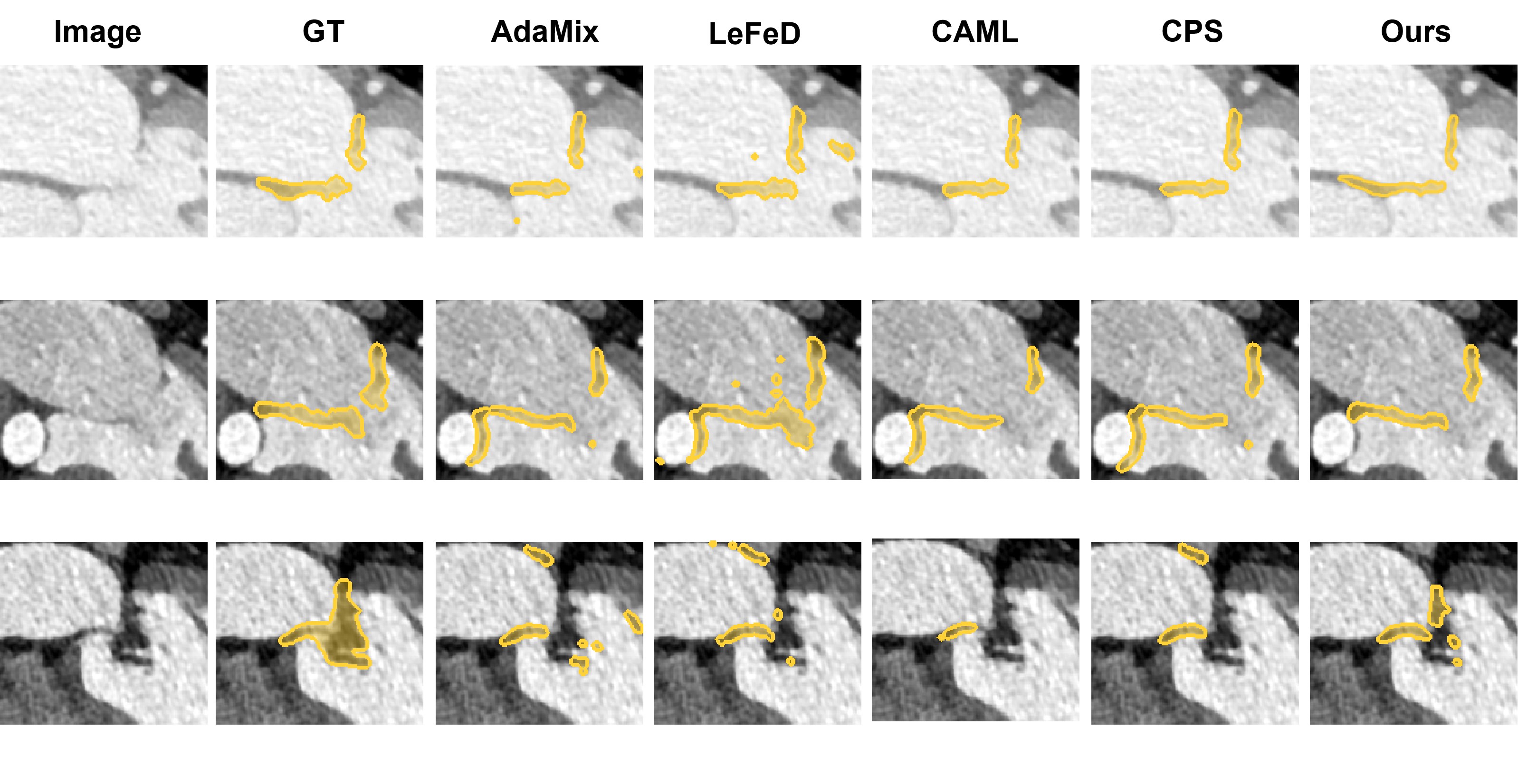}
    \caption{Visual comparison on representative slices from the held-out MVAA-CT test split. Yellow contours denote the annotated segmentation regions.}
    \label{fig:mvaa_visual_comparison}
\end{figure}

Figure~\ref{fig:acdc_visual_comparison} and Figure~\ref{fig:mvaa_visual_comparison} show visual comparisons on ACDC and MVAA-CT, respectively. On ACDC, ProGeo-RC produced more stable segmentation around myocardial boundaries, ventricular contours, and low-contrast regions than AdaMix~\cite{b30}, BCP~\cite{b13}, CPS~\cite{b12}, and MC-Net+~\cite{b27}, with better preservation of RV, Myo, and LV shapes and fewer boundary shifts. On MVAA-CT, where thin valve structures are prone to discontinuities, boundary shifts, and background false positives, ProGeo-RC showed more accurate localization, better structural continuity, and fewer isolated false-positive regions. These visual observations are consistent with the quantitative results.

\begin{table}[!t]
\centering
\caption{Analysis of model parameters and computational complexity on ACDC with an input size of $256\times256$.}
\label{tab:efficiency_acdc}
\scriptsize
\setlength{\tabcolsep}{10pt}
\renewcommand{\arraystretch}{1.02}
\begin{tabular}{lcc}
\toprule
Method & Params (M) & GFLOPs \\
\midrule
AdaMix   & 3.27 & 7.40 \\
BCP      & 1.81 & 6.01 \\
CPS      & 1.81 & 6.01 \\
DiffRect & 1.81 & 6.01 \\
MC-Net   & 3.6 & 14.02 \\
MC-Net+  & 3.5 & 13.81 \\
MLRP     & 2.45 & 9.85 \\
\textbf{Ours} & \textbf{1.81} & \textbf{6.01} \\
\bottomrule
\end{tabular}
\end{table}

\subsection{Computational Efficiency}
\label{subsec:efficiency}
Computational efficiency is important for clinical deployment. As shown in Table~\ref{tab:efficiency_acdc}, ProGeo-RC requires 1.81M parameters and 6.01 GFLOPs under single-model inference, matching BCP, CPS, and DiffRect while using fewer resources than AdaMix, MC-Net, MC-Net+, and MLRP. Since PSRE and S-SGEC act only as training-stage pseudo-supervision calibration strategies, they introduce no additional inference-time branches. Thus, ProGeo-RC improves segmentation performance while preserving the inference complexity of the underlying backbone.

\section{Conclusion}
\label{sec:conclusion}
We presented ProGeo-RC, a semi-supervised cardiac image segmentation framework that improves pseudo-supervision by coupling feature-space semantic reliability with image-space geometric plausibility. PSRE estimates reliable unlabeled pixels and calibrates teacher pseudo labels using probabilistic prototypes, while S-SGEC applies signed-distance-field-based constraints to refine ambiguous boundaries and suppress abnormal foreground responses. Experiments on ACDC and MVAA-CT, together with ablation studies and visualizations, demonstrated consistent gains across annotation settings, imaging modalities, and segmentation targets. The efficiency analysis further showed that ProGeo-RC introduces no extra inference-time architectural overhead. These results indicate that joint semantic and geometric calibration can provide reliable and efficient pseudo-supervision for low-annotation cardiac image analysis.

\end{document}